\documentclass[runningheads]{llncs}

\usepackage{multirow}
\usepackage{booktabs}
\usepackage{amsmath}
\usepackage{amssymb}
\usepackage{graphicx}
\usepackage{arydshln}
\usepackage{caption}
\usepackage[table]{xcolor}
\definecolor{bestcolor}{HTML}{bce6cd}
\definecolor{secondcolor}{HTML}{e4eebc}
\definecolor{thirdcolor}{HTML}{fef8c4}

\newcommand{\bestbg}[1]{\cellcolor{bestcolor}\textbf{#1}}
\newcommand{\secondbg}[1]{\cellcolor{secondcolor}#1}
\newcommand{\thirdbg}[1]{\cellcolor{thirdcolor}#1}
\usepackage{subcaption}
\usepackage{cite}
\usepackage{hyperref}
\hypersetup{hidelinks}
\usepackage[capitalise,nameinlink]{cleveref}
\usepackage{orcidlink}

\begin{document}


\title{Real-Time LiDAR Gaussian Splatting SLAM
\\
via Geometry-Aware Covariance Coupling}

\titlerunning{Real-Time LiDAR GS SLAM}

\author{SeungJun Tak\thanks{These authors contributed equally to this work.}\orcidlink{0009-0008-2204-6137} \and
Yewon Jeon\protect\footnotemark[1]\orcidlink{0009-0003-6715-5445} \and
Jaeik Hwang\orcidlink{0009-0006-3047-0320} \and
Suk Min Hwang\orcidlink{0009-0008-1039-9074} \and
Seongbo~Ha\orcidlink{0009-0007-7018-1598} \and
Hyeonwoo Yu\orcidlink{0000-0002-9505-7581}
}

\authorrunning{SeungJun Tak et al.}

\institute{Sungkyunkwan University, Suwon, South Korea\\
\email{\{tmdwns8840,yw8431,jaeik0211,raph03,sobo3607,hwyu\}@skku.edu}
}
\maketitle

\begin{abstract}
We present a real-time LiDAR-based framework for Gaussian Splatting SLAM that tightly couples fast G-ICP registration with spherical rasterization-based dense mapping for large-scale sequences. 
Leveraging LiDAR geometry rather than appearance, we reuse tracking-estimated local covariances to initialize Gaussians with range-aware scales and to derive surface normals for geometry-aware map optimization. 
We further introduce a covariance-derived geometry score that measures local complexity and drives pruning in planar regions and selective densification in structurally rich areas, while optimized Gaussians and LiDAR-specific confidence cues are fed back to improve tracking robustness. 
On the Newer College dataset, our method achieves an F-score of 86.78\% using purely online trajectories at real-time speed ($>$20 FPS), and additional experiments on other datasets confirm its stability and scalability.
\\
Code is available at
\href{https://github.com/Lab-of-AI-and-Robotics/LiDAR-GS-SLAM}{github.com/Lab-of-AI-and-Robotics/LiDAR-GS-SLAM}.
\\
Project Page:
\href{https://lab-of-ai-and-robotics.github.io/GS-SLAM-Family}{lab-of-ai-and-robotics.github.io/GS-SLAM-Family}

\end{abstract}

\begin{figure*}[t]
    \centering
    \includegraphics[width=\textwidth]{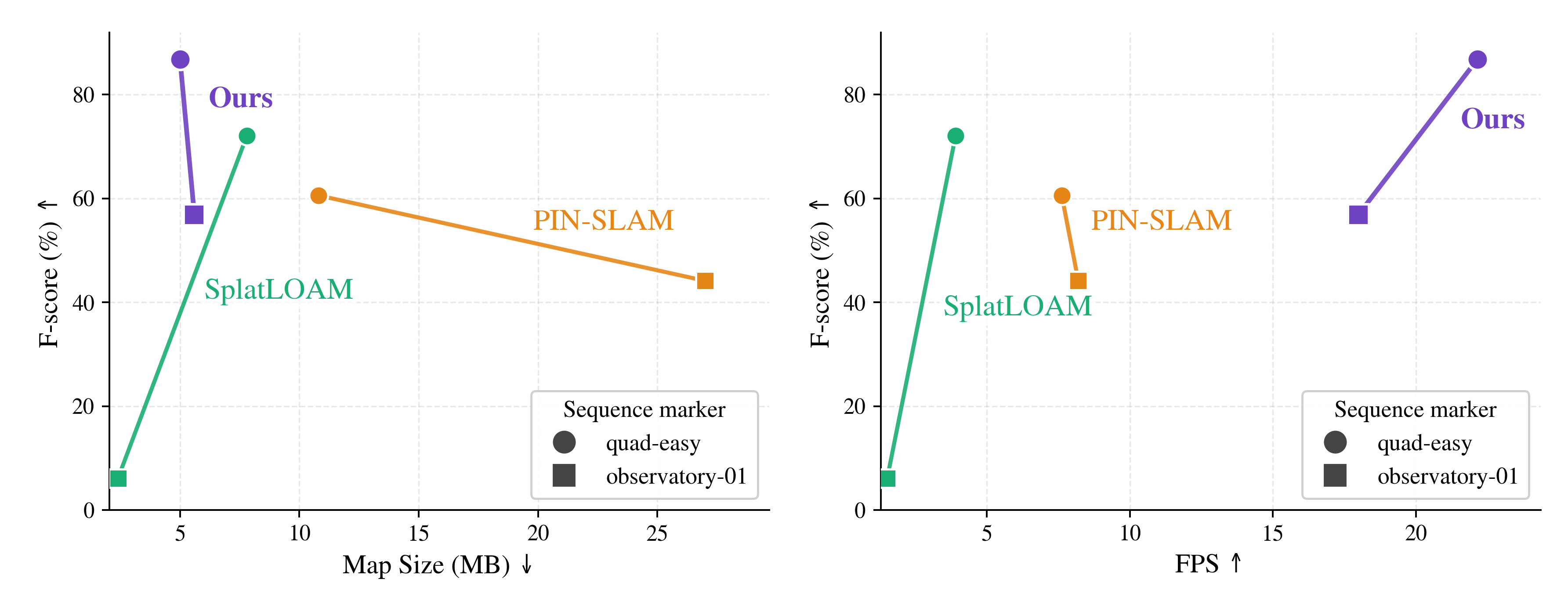}
     
    \caption{\textbf{Performance Overview of LiDAR GS.} We compare dense mapping SLAM with respect to Map Size (MB) and Frame per Second (FPS), with F-score (\%). The two panels illustrate the trade-off between efficiency and accuracy.}
    \label{fig:Performance_overview}
     
\end{figure*}

\section{Introduction}

LiDAR-based SLAM~\cite{zhang2014loam,shan2020lio,xu2021fast,kiss2025iros} provides robust trajectory tracking in large-scale environments, but most systems rely on sparse point maps or voxel structures for real-time operation. While effective for localization, these representations are limited for high-fidelity reconstruction and visibility-aware reasoning, which are increasingly demanded in autonomous-driving simulation and HD map production.
Recently, 3D Gaussian Splatting (3DGS)~\cite{kerbl3Dgaussians} has emerged as an efficient dense representation, yet integrating it into online large-scale SLAM remains challenging due to (i) \emph{unbounded map growth} that inflates GPU memory and rendering/registration cost, and (ii) \emph{error persistence}, where corrupted primitives bias subsequent registrations over long horizons.

We address these issues with a LiDAR-based Gaussian Splatting SLAM system that enables scalable and reliable dense mapping in large-scale outdoor environments. We perform fast tracking using Generalized ICP~\cite{segal2009generalized} against a trackable subset of the map. For mapping, we adopt rasterization-based optimization but represent primitives as surface-oriented 2D anisotropic Gaussians~\cite{huang20242d} in spherical coordinates, which better matches LiDAR surface observations.
Crucially, we tightly couple tracking and mapping by reusing geometric information across stages with minimal overhead: tracking covariances provide geometric cues for Gaussian initialization and online map control, while mapping-refined geometry reweights tracking residuals for robust registration under limited observations.

This coupling is realized through two mechanisms. First, we derive normals and anisotropic scales directly from local G-ICP covariances and store them as primitive parameters, stabilizing and accelerating map optimization without additional learning or separate geometry estimation. Second, we enforce scalability via geometry-aware online map management, performing targeted pruning in planar/redundant regions and selective densification in structurally informative areas. We further infer a reliability mask from ray-drop patterns and observation-induced normal inconsistencies, and use it consistently in both mapping (to suppress unreliable primitives) and tracking (for correspondence filtering/weighting) to improve long-term robustness.

We evaluate our method on large-scale LiDAR benchmarks including KITTI Odometry~\cite{Geiger2012CVPR}, Oxford Spires~\cite{tao2025spires}, and Newer College~\cite{ramezani2020newer}, comparing against classical and neural field-based LiDAR SLAM baselines. Results on trajectory accuracy (ATE RMSE), runtime (FPS), and map growth (number of primitives) demonstrate that our system achieves competitive accuracy while substantially improving scalability and reliability. Our main contributions are as follows:
\begin{itemize}
\item \textbf{System:} We propose a real-time scalable LiDAR Gaussian Splatting SLAM system that builds dense, continuous maps in large-scale outdoor scenes.
\item \textbf{Mutual reuse of covariance-derived geometric cues:} 
Tracking covariances initialize Gaussian geometry for mapping, while mapping-refined Gaussians provide covariance-aware targets for robust registration.
\item \textbf{Reliability \& map management:} We present geometry-aware pruning/den\-sification and an observation-driven reliability mask, jointly used in tracking and mapping for robust long-horizon operation.
\end{itemize}

\section{Related Work}
\textbf{LiDAR Odometry and SLAM. }\quad

To achieve real-time processing, traditional large-scale LiDAR SLAM systems abstract scenes into sparse point sets or low-resolution voxels. The field has evolved from early curvature-based feature extraction~\cite{zhang2014loam, shan2018lego} to tightly-coupled optimization frameworks~\cite{shan2020lio, xu2022fast, he2023point} and continuous-time trajectory modeling~\cite{dellenbach2022ct, chen2022direct, kiss2025iros} for enhanced robustness against motion and noise. However, since these methods primarily target odometry and localization, they maintain sparse map structures tailored for registration, neglecting explicit continuous surface representations or visibility-aware operations.
\newline
\textbf{Dense Mapping and Implicit Representation. }\quad
To overcome sparse maps, dense mapping has modeled surface continuity, from volumetric TSDF~\cite{oleynikova2017voxblox, vizzo2022vdbfusion} and surfel maps~\cite{behley2018efficient,chen2019suma++} to neural/implicit representations~\cite{deng2023nerf, zhong2023shine, yang2022vox, turki2022mega, pan2024pin} that enable high-quality reconstruction. 
However, in online large-scale SLAM, frequent updates and global re-optimization incur high cost and hinder real time. Over long trajectories, accumulated noise and drift also make map consistency difficult, degrading tracking stability. 
Thus, balancing geometric fidelity and online scalability remains open.
\newline

\textbf{Gaussian Splatting-based SLAM. }\quad
Recently, Gaussian Splatting~\cite{kerbl3Dgaussians, huang20242d} has emerged as an effective representation that enables real-time rendering with explicit spatial extents. 
Early vision-based GS-SLAM studies~\cite{keetha2024splatam, yan2024gs, matsuki2024gaussian, huang2024photo} optimized Gaussians mainly with dense photometric cues, while subsequent works coupled G-ICP with Gaussian mapping for RGB-D~\cite{ha2024rgbd,pak2025g2s}, introduced bidirectional monocular VO--Gaussian coupling~\cite{yeon2026gso}, and explored large-scale outdoor operation with LiDAR--camera--IMU fusion~\cite{xiao2024liv, lang2025gaussian, xie2025gs, liu2025gs}.
LiDAR-only dense Gaussian mapping remains more challenging due to sparse range observations and the lack of photometric cues. Splat-LOAM~\cite{giacomini2025splat} uses spherical LiDAR projection for geometry-only reconstruction, while NeRF-LOAM~\cite{deng2023nerf} is LiDAR-only but much slower in our online evaluation. In contrast to RGB-D or LiDAR-visual systems such as GS-ICP, G2S-ICP~\cite{ha2024rgbd,pak2025g2s}, PINGS\cite{pan2025pings}, and GSO-SLAM\cite{yeon2026gso}, our protocol is LiDAR-only, geometry-centered, and online.

\section{Method}

\begin{figure*}[t]
    \centering
    \includegraphics[width=\textwidth]{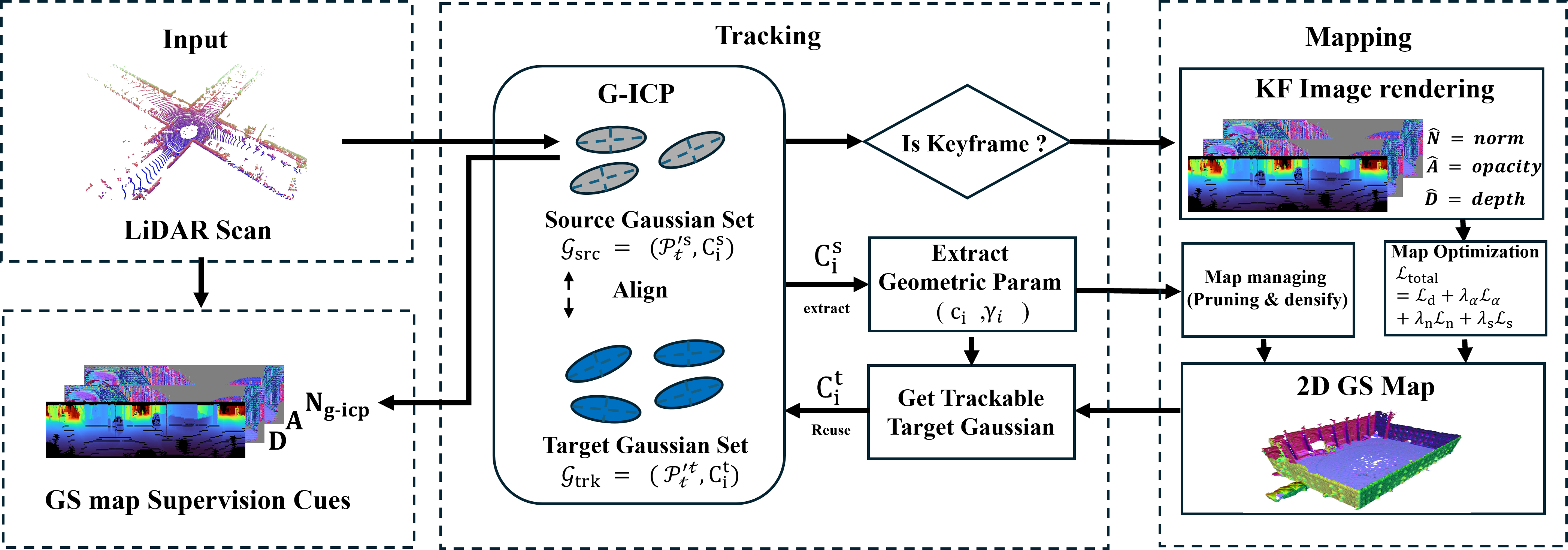}
    \caption{\textbf{Overview of the proposed system.} We downsample each LiDAR scan and estimate per-point covariances to form the source set $\mathcal{P}'^{\,s}_{t}$. Tracking registers $\mathcal{P}'^{\,s}_{t}$ to a trackable target set from the map via G-ICP to estimate $T_t$, and the covariances produce a control score $c_i$ for pruning/densification. Keyframes are fused into the 2D Gaussian map and optimized, while reusing stored target parameters avoids per-frame covariance re-estimation and improves robustness to geometric noise.}
    \label{fig:system_overview}
\end{figure*}

In this paper, we propose a real-time LiDAR Gaussian Splatting SLAM system for large-scale outdoor environments. We normalize LiDAR scans to the spherical domain and represent the map as 2D Gaussian primitives rasterized in the same domain. While each Gaussian stores rich parameters, memory usage and optimization cost can grow rapidly as the map scales. To address this, we reuse the registration residuals and local geometric cues already computed in G-ICP~\cite{segal2009generalized} tracking. Without an extra geometric estimation module, these signals initialize new primitives, set their geometric parameters, and guide densification and pruning. Conversely, geometric parameters and uncertainty refined during mapping are fed back to tracking to form a trackable subset and perform covariance-weighted G-ICP, reducing correspondence search cost and mitigating error accumulation over long trajectories. The overall pipeline is summarized in Fig.~\ref{fig:system_overview}, after which we present the spherical-domain 2D Gaussian map, trackable Gaussian selection with geometry-weighted G-ICP tracking, tracking-informed priors for mapping, 2DGS optimization, and geometry-aware map management.
\subsection{Map Representation}
\label{sec:2dgs}

We represent the map as 2D Gaussian primitives
$\mathcal{G}=\{(\boldsymbol{\mu}_i,\mathbf{q}_i,\tilde{\mathbf{s}}_i,\alpha_i)\}_{i=1}^{N}$,
where $\boldsymbol{\mu}_i\in\mathbb{R}^3$ is the mean, $\mathbf{q}_i$ is the orientation,
$\tilde{\mathbf{s}}_i=[s_{i,0},s_{i,1}]$ is the in-plane scale, and $\alpha_i$ is the opacity.
Each primitive additionally stores per-Gaussian signals $(u_i,c_i)$ for LiDAR confidence and map complexity control.
For mapping, we build each scan as a spherical range image and render the Gaussian map under the same spherical projection
$\pi_s(\cdot)$, producing $(\hat D,\hat A,\hat{\mathbf N})$ on a shared grid for stable geometric supervision.

\subsection{G-ICP Tracking with Trackable Gaussians}
\label{sec:tracking}
Given an incoming scan at time $t$, we form the source set $\mathcal{P}^s_t=\{\mathbf{p}^s_i\}$
and apply voxel-grid downsampling (voxel size $\delta$) to obtain $\mathcal{P}'^{\,s}_{t}$.
To reduce correspondence cost and avoid stale/unstable map elements, we register only against a trackable subset:
\begin{equation}
\mathcal{G}_{\text{trk}}=\{g_i\in\mathcal{G}\mid m_i=1,\;k_i\ge k_{\min},\;\alpha_i\ge \alpha_{\min}\},
\label{eq:trackable_select}
\end{equation}
where $m_i$ is a trackable mask and $k_i$ denotes the last keyframe index that updated $g_i$.
We convert $\mathcal{G}_{\text{trk}}$ into a target point set $\mathcal{P}'^{\,t}_t$ by taking Gaussian means
(and optional voxel downsampling for efficiency).
Using $\mathcal{P}'^{\,s}_t$ and $\mathcal{P}'^{\,t}_t$, we estimate $\mathbf{T}_t\in SE(3)$ via G-ICP.

\noindent\textbf{Covariance construction and reuse. }
During tracking, G-ICP estimates a local covariance for each downsampled source point
$\mathbf{p}'^{\,s}_i\in\mathcal{P}'^{\,s}_{t}$ from its $k$-NN neighborhood
$\mathcal{N}(\mathbf{p}'^{\,s}_i)$.
We cache its eigendecomposition (principal axes and standard deviations),
$(\mathbf{R}^s_i,\boldsymbol{\sigma}_i)$, which is already computed inside the tracker.
When a keyframe is inserted, these source covariances are reused as low-cost priors to initialize
the rotation, anisotropic in-plane scale, and confidence of newly created map Gaussians.

Conversely, for tracking we construct target covariances directly from each map Gaussian $g_i$:
its orientation $\mathbf{q}_i$ and in-plane scales $(s_{i,x},s_{i,y})$ define a surfel-like $3{\times}3$
covariance with a small thickness $\epsilon$ along the normal direction.
This enables $O(1)$ target covariance construction without any map-side $k$-NN search.
Using these target covariances, the tracker performs covariance-weighted G-ICP between
$\mathcal{P}'^{\,s}_{t}$ and the trackable target set $\mathcal{P}'^{\,t}$ to estimate
$\mathbf{T}_t\in SE(3)$.
As mapping optimizes $(\mathbf{q}_i, s_{i,x}, s_{i,y})$, the target covariances become more accurate,
which in turn stabilizes subsequent registrations. For the very first frame (before a map exists),
we fall back to the same neighborhood-based covariance estimation used for sources.

\noindent\textbf{Geometry-aware Matching Weight. }
We leverage planar priors to improve LiDAR tracking~\cite{zhou2021pi,7989597}.
Given the in-plane Gaussian scales $s_i=[s_{i,0},s_{i,1}]~(s_{i,0}>s_{i,1})$, we define an curvature
\begin{equation}
\kappa_i = \frac{s_{i,0}}{s_{i,1}+\epsilon},
\label{eq:anisotropy}
\end{equation}
and normalize it to $\bar\kappa_i\in[0,1]$:
\begin{equation}
\bar\kappa_i=\frac{\kappa_i-\kappa_{\min}}{\kappa_{\max}-\kappa_{\min}}.
\label{eq:anisotropy_norm}
\end{equation}
Lower $\bar\kappa_i$ indicates a more planar primitive. We then solve a weighted G-ICP objective:
\begin{equation}
\mathbf{T}_t = \arg\min_{\mathbf{T}} \sum_{i} w(\bar\kappa_i)\;
\mathbf{d}_i^\top \left( \mathbf{C}^t_i + \mathbf{T} \mathbf{C}^s_i \mathbf{T}^\top \right)^{-1} \mathbf{d}_i,
\label{eq:gicp_weighted}
\end{equation}
with
\begin{equation}
w(\bar\kappa_i)=\min\left(\frac{1}{\bar\kappa_i+\epsilon},\omega_{\max}\right),\quad \omega_{\max}=10.
\label{eq:weight_func}
\end{equation}
This upweights stable planar structures, acting as geometric anchors and reducing orientation drift.\\
\noindent\textbf{Loop Closure. }
In addition to local tracking, we leverage loop-closure to reduce long-term drift. 

Upon detecting a loop~\cite{gupta2024icra}, we compute the relative pose between the source keyframe and the matched target keyframe by aligning the source keyframe with its corresponding submap. This relative pose is then introduced as a new constraint in the pose graph. Following pose-graph optimization, each Gaussian is updated by the optimized-pose delta of its source keyframe, yielding a keyframe-wise piecewise-rigid map correction rather than full global Gaussian re-optimization.
%
\subsection{Tracking-Derived Initialization and Geometric Parameters}
\label{sec:ttm} 

Mapping optimization is highly sensitive to initialization, strongly impacting stability and convergence. 
In large-scale settings, per-point neighbor searches or extra geometric estimation are prohibitively expensive. 
We therefore reuse local geometric statistics already produced by G-ICP tracking to initialize new Gaussians and derive low-cost map-management signals, achieving stable convergence with real-time efficiency. 
Because LiDAR provides limited appearance cues, these geometric statistics are crucial for preserving map quality and stable tracking. 
This section details covariance-based Gaussian initialization and the two management signals: control score and trackability confidence.\\
%
\noindent\textbf{Range-adaptive Scale Initialization. }
\label{sec:scale_init}
G-ICP tracking provides a covariance decomposition summarized by the standard-deviation vector
$\boldsymbol{\sigma}_i=[\sigma_{i,0},\sigma_{i,1},\sigma_{i,2}]$, sorted as
$\sigma_{i,0}\ge\sigma_{i,1}\ge\sigma_{i,2}$.
We use the two largest components as tangent-plane standard deviations
$\boldsymbol{\sigma}^{\text{tan}}_i=[\sigma_{i,0},\sigma_{i,1}]$.

Because LiDAR has fixed angular resolution, surface sampling becomes sparser with range $r$.
However, G-ICP typically normalizes covariances for numerical stability, so
$\boldsymbol{\sigma}^{\text{tan}}_i$ mainly reflects local anisotropy rather than absolute spacing.
To ensure sufficient initial coverage at long range, we initialize the 2D scale proportional to $r$:
for $r_i=\max(\|\mathbf{p}_i\|,1)$,
\begin{equation}
\mathbf{s}_i^{(0)}
=
\mathrm{clip}\!\Big(
\kappa \, r_i \, \boldsymbol{\sigma}^{\text{tan}}_i,\,
0,\,
s_{\max}
\Big),
\end{equation}
and optimize the log-scale $\tilde{\mathbf{s}}_i=\log(\mathbf{s}_i)$.
This range-adaptive initialization reduces far-range holes and stabilizes subsequent map optimization.\\
\noindent\textbf{LiDAR-specific Opacity and Trackability Confidence. }
\label{sec:opacity}
Unlike image-based Gaussian splatting where opacity is optimized to explain photometric rendering, LiDAR returns provide only sparse geometry and their reliability is dominated by sensing geometry.
Therefore, we reinterpret opacity as a LiDAR-specific reliability weight and initialize it from the physics-based confidence $u$, making it well aligned with our sensor for tracking and map management.
LiDAR measurements vary in reliability depending on range and incidence angle.
Given each point range $r=\|\mathbf{p}\|$, the normal $\mathbf{n}$, and the sensor ray direction $\hat{\mathbf{r}}=\mathbf{p}/\|\mathbf{p}\|$,
we compute the incidence cosine term $c=|\mathbf{n}^\top\hat{\mathbf{r}}|$ and define the physics-based confidence $u$ as
\begin{equation}
u
=
\exp\!\Big(-\big(\tfrac{r}{r_0}\big)^2\Big)\cdot
\mathrm{clip}\!\Big(\tfrac{c-c_0}{1-c_0+\epsilon},\,0,\,1\Big).
\label{eq:physconf}
\end{equation}
When adding a new Gaussian in mapping, points with larger $u_i$ are initialized with higher opacity:
\begin{equation}
\alpha_i^{(0)}=\alpha_{\min}+(\alpha_{\max}-\alpha_{\min})\,u_i.
\label{eq:alpha_init}
\end{equation}
We further define the trackability confidence as $\gamma_i = \alpha_i \, u_i$,
which combines learned opacity and physics-based reliability.

\noindent\textbf{Control Score. }
\label{sec:controlscore}
In large-scale sequences, Gaussians accumulate over time, causing memory and rendering costs to grow roughly linearly. To keep the map scalable, we reallocate resources by compressing planar regions and selectively densifying complex structures. To this end, we define a control score $c_i$ from covariance-based features computed during tracking. Using the covariance eigenvalues $\lambda_{i,j}=\sigma_{i,j}^2$, we compute linearity and curvature as
\[
\mathrm{linear}_i=\frac{\lambda_{i,0}-\lambda_{i,1}}{\lambda_{i,0}+\epsilon},\qquad
\mathrm{curv}_i=\frac{\lambda_{i,2}}{\lambda_{i,0}+\lambda_{i,1}+\lambda_{i,2}+\epsilon}.
\]
We also include the normalized G-ICP registration residual $\widetilde{\mathrm{res}}_i$ and define the control score as a
linear combination of normalized terms:
\begin{equation}
c_i = \mathrm{clip}\!\Big(
w_l \widetilde{\mathrm{linear}}_i +
w_c \widetilde{\mathrm{curv}}_i +
w_r \widetilde{\mathrm{res}}_i,\;
0,\;1\Big).
\label{eq:controlscore}
\end{equation}
In our implementation, we use $w_l=0.55,\; w_c=0.30,\; w_r=0.15$.
A lower $c_i$ indicates more planar structure, while a higher $c_i$ indicates more geometrically complex structures (e.g., edges/corners).
This score is used for (i) planar compression (cover-and-prune) and (ii) selective splitting of complex regions (Section~\ref{sec:mapmanage}).
%
\subsection{Map Optimization}
\label{sec:mapping}
Mapping rasterizes the current Gaussian map into the range-image domain to obtain predicted depth, opacity, and normals $(\hat{D},\hat{A},\hat{\mathbf{N}})$, and optimizes the Gaussian parameters
$\Theta=\{\boldsymbol{\mu},\mathbf{q},\tilde{\mathbf{s}},\alpha\}$ by comparing them against the observations.
To avoid invalid measurements, we optimize only over valid pixels using depth-valid mask $\mathbf{M}_D=\mathbb{I}[D>0]$.

\noindent\textbf{Depth \& Opacity Loss. }
Depth and opacity losses are defined on valid LiDAR pixels using the valid-depth mask $\mathbf{M}_D$.
Specifically, we use a masked L1 loss for depth and a positive-only BCE (i.e., $\mathrm{BCE}(\hat{A},\mathbf{1})$) for opacity:
\begin{equation}
\mathcal{L}_{\text{depth}}
=
\frac{1}{|\Omega_D|}
\sum_{i\in\Omega}
\mathbf{M}_{D,i}\, \big|\hat{D}_i - D_i\big|,
\quad
\mathcal{L}_{\alpha}
=
-
\frac{1}{|\Omega_D|}
\sum_{i\in\Omega}
\mathbf{M}_{D,i}\log(\hat{A}_i+\epsilon),
\label{eq:depthloss}
\end{equation}

where $\Omega$ denotes the set of all pixels and $\Omega_D=\{i\in\Omega \mid \mathbf{M}_{D,i}=1\}$.

\noindent\textbf{Normal Loss with Valid Masks. }
Normal alignment complements the weak geometric constraints of depth-only supervision, stabilizing surface orientation and boundary learning.
We employ two types of normals: (i) geometric normals $\mathbf{N}_{\text{g-icp}}$ obtained from tracking (G-ICP), and
(ii) surface normals $\mathbf{N}_{\text{surf}}$ computed from the rendered depth.
$\mathbf{N}_{\text{g-icp}}$ is strong for local planar estimation but may be unstable in some regions, while
$\mathbf{N}_{\text{surf}}$ provides observation-driven boundary/discontinuity cues; thus they are complementary.
To avoid early-stage divergence, we activate the $\mathbf{N}_{\text{surf}}$ term only after densification via a scheduling strategy.
We define the valid mask for G-ICP normals as
\[
\mathbf{M}_N = \mathbf{M}_D \odot \mathbb{I}[\|\mathbf{N}_{\text{g-icp}}\|>0],
\]
and use the cosine loss $\mathcal{L}_{\cos}(\mathbf{a},\mathbf{b})=1-\mathbf{a}^\top\mathbf{b}$:
\begin{equation}
\mathcal{L}_{n}
=
\lambda_{n,g}\,
\frac{1}{|\Omega_N|}
\sum_{i\in\Omega_N}
\mathcal{L}_{\cos}\big(\hat{\mathbf{N}}_i,\mathbf{N}_{\text{g-icp},i}\big)
+
\lambda_{n,s}\,
\frac{1}{|\Omega_D|}
\sum_{i\in\Omega_D}
\mathcal{L}_{\cos}\big(\hat{\mathbf{N}}_i,\mathbf{N}_{\text{surf},i}\big),
\label{eq:normalloss}
\end{equation}
where $\Omega_N=\{i\in\Omega \mid \mathbf{M}_{N,i}=1\}$.
For training stability, $\lambda_{n,s}$ is enabled only after densification begins.

\noindent\textbf{Scale Regularization Loss. }
To prevent abnormally large Gaussians from causing geometric distortions, we penalize only when the maximum component of the 2D scale exceeds a threshold $s_{\max}$:
\begin{equation}
\mathcal{L}_{\text{scale}}
=
\sum_{i=1}^{N}
\max\big(0,\; \max(s_{i,x},s_{i,y})-s_{\max}\big).
\label{eq:scaleloss}
\end{equation}
%
The final mapping optimization minimizes the following combined objective:
\begin{equation}
\mathcal{L}
=
\mathcal{L}_{\text{depth}}
+
\lambda_{\alpha}\mathcal{L}_{\alpha}
+
\lambda_{n}\mathcal{L}_{n}
+
\lambda_{s}\mathcal{L}_{\text{scale}}.
\label{eq:totalloss}
\end{equation}
\subsection{Real-time Map Budget Control for Long Sequences}
\label{sec:mapmanage}

\begin{figure*}[t]
    \centering
    \includegraphics[width=\textwidth]{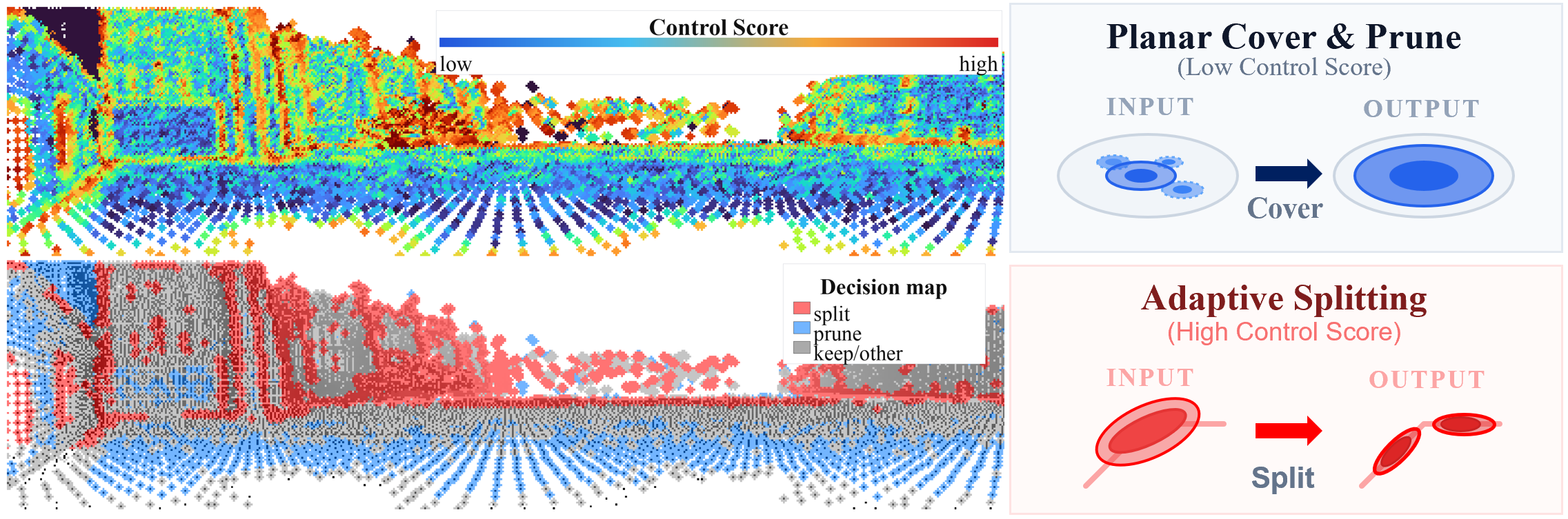}
    
    \caption{\textbf{Covariance-derived control score and decision map. }
From each tracking covariance, we compute linearity/curvature proxies from eigenvalues and fuse them with a robustly normalized G-ICP residual to obtain a control score $c_i\in[0,1]$.
Low $c_i$ indicates planar/redundant regions and triggers cover-and-prune, while high $c_i$ indicates geometric complexity (edges/corners) and triggers selective splitting.
The visualization shows how decisions spatially correlate with scene structure.}
    \label{fig:controlscore_decisionmap}
    
\end{figure*}
Large-scale sequences continuously accumulate keyframes, leading to steady Gaussian growth and increasing memory, rendering, and optimization costs that threaten real-time operation. Beyond opacity-based pruning/densification, we therefore introduce a tracking-derived control score $c_i$, enabling an efficient map representation under long sequences.

\noindent\textbf{Multi-pass Planar Cover and Prune. }
To efficiently reduce redundant Gaussians on large planar structures, we perform a \textbf{multi-pass} grouping strategy based on projections onto orthogonal planes, instead of relying on a single 3D voxelization. In each pass, we project each Gaussian mean $\mu_i$ onto the corresponding plane and form voxel-grid groups $G$ in projected 2D domain.
For each group $G$, we consider only Gaussians that satisfy (i) a low control score $c_i \leq Q_{q_{plane}}(c)$ and (ii) sufficient opacity $\alpha_{min}\leq \alpha_i$. We then select the Gaussian with the largest $\alpha$ as the representative $g_r$, and expand its in-plane-scale in the local tangent-plane coordinates so that the cover region includes a $q_{cover}$. Among the members falling inside the cover region, planar candidates are treated as redundant and removed. To prevent excessive information loss, we cap the maximum pruning ratio per pass.

\noindent\textbf{Structure-aware Adaptive Splitting. }
A high control score $c_i$ indicates structurally complex regions such as edges and corners. To selectively enhance representational capacity in these areas, we perform adaptive splitting. At each map-management step, we select split candidates whose control scores are above the $q_{split}$, while also requiring $\alpha_{min}\leq\alpha_i$ to avoid over-splitting unstable primitives. We further limit resource growth by capping the splitting ratio relative to the total number of Gaussians N.
For each split candidate $g_i$, we define the in-plane major axis as $a_i = R(q_i)e_0$ and generate two child Gaussians by shifting the mean in opposite directions along $a_i$. This symmetry-breaking initialization causes the two children to receive different gradients during subsequent optimization, allowing them to specialize to nearby geometric variations and  represent fine structures.
\cref{fig:controlscore_decisionmap} visually demonstrates how the decisions derived from these techniques align with the actual scene structure.

\section{Experiments}

\subsection{Experimental Setup}
\noindent\textbf{Datasets. }

We evaluate the proposed method on the KITTI Odometry dataset~\cite{Geiger2012CVPR}, the Oxford Spires dataset~\cite{tao2025spires}, and the Newer College dataset~\cite{ramezani2020newer}.
KITTI contains large-scale long-horizon sequences in autonomous driving scenarios, and is thus used to assess tracking stability and scalability in large-scale environments.
In contrast, Newer College and Oxford Spires are captured with a handheld sensor at relatively low speed, enabling dense observations; moreover, they provide ground-truth meshes, which allow us to evaluate both map quality and local registration stability.

For every dataset, we deskew each scan by interpolating the estimated ego-motion over time.

\noindent\textbf{Implementation Details. }

All experiments are conducted on a desktop computer equipped with an AMD Ryzen 9 7900X CPU, 64 GB of RAM, and an NVIDIA GeForce RTX 4090 GPU. For the mapping loss (Eqn.~\ref{eq:totalloss}), the weights are set to $\lambda_{\alpha}=0.1$, $\lambda_n=1$, and $\lambda_s=10$. For Eqn.~\ref{eq:normalloss}, we set $\lambda_{n,g}=0.05$ and $\lambda_{n,s}=0.01$.

\noindent\textbf{Baselines. }

We compare the proposed method with various map representation approaches, including point~\cite{kiss2025iros}, surfel~\cite{behley2018efficient}, TSDF~\cite{oleynikova2017voxblox, vizzo2022vdbfusion}, neural field~\cite{pan2024pin, song2024n}, and Gaussian Splatting~\cite{giacomini2025splat}.
We evaluate our approach against mapping-only methods~\cite{oleynikova2017voxblox, vizzo2022vdbfusion, song2024n}. 
Since these methods operate using ground truth poses, they are free from the mapping degradation typically caused by pose estimation errors. Consequently, they represent an ideal scenario and serve as an upper bound for the achievable map quality.

\noindent\textbf{Metrics. }

To evaluate tracking accuracy, we employ the Root Mean Square Error (RMSE) of the Absolute Trajectory Error (ATE). For the mapping evaluation, we assess mesh-to-mesh quality using Accuracy (Acc), Completeness (Com), Chamfer-L1 (C-L1), and the $\tau$-F-score (\%), which are based on bidirectional nearest-neighbor distances. All distance metrics are reported in centimeters (cm). In our implementation, we set $\tau=0.2$ m, the downsampling resolution to 0.02 m, and truncate distance outliers at 0.5 m.
\begin{table*}[t]
\centering
\caption{
\textbf{Trajectory evaluation results (ATE RMSE [m] $\downarrow$) on Newer College, Oxford Spires, and KITTI datasets.}
Tracking accuracy is measured using the ATE RMSE in meters. The best, second-best, and third-best results are highlighted in \bestbg{first}, \secondbg{second}, and \thirdbg{third} colors, respectively. `Fail' indicates that the method lost tracking or failed to converge on the corresponding sequence.
}
\label{tab:total_ate_evaluation}
\resizebox{0.75\textwidth}{!}{
\begin{tabular}{lccccccccc}
    \toprule
    \textbf{Method} & \multicolumn{2}{c}{\textbf{Newer}} & \multicolumn{2}{c}{\textbf{Oxford}} & \multicolumn{5}{c}{\textbf{KITTI}} \\
    \cmidrule(lr){2-3} \cmidrule(lr){4-5} \cmidrule(lr){6-10}
    & {quad} & {math} & {obs} & {keb} & {00} & {01} & {07} & {08} & {09} \\
    \midrule
    KISS-SLAM~\cite{kiss2025iros} & \thirdbg{0.100} & \secondbg{0.151} & \thirdbg{0.494} & \secondbg{0.367} & 1.699 & \thirdbg{19.632} & 0.435 & 3.982 & \thirdbg{1.687} \\
    PIN-SLAM~\cite{pan2024pin}  & \secondbg{0.097} & \bestbg{0.071} & \bestbg{0.150} & 12.196 & \bestbg{0.913} & \secondbg{3.670} & \bestbg{0.281} & \bestbg{1.778} & \bestbg{1.246} \\
    SuMa~\cite{behley2018efficient}      & 0.249 & 0.210 & 5.145 & 16.803 & \thirdbg{1.200} & Fail & 0.405 & \secondbg{2.210} & 3.884 \\
    Splat-LOAM~\cite{giacomini2025splat} & 0.138 & 0.818 & 17.494 & \thirdbg{0.770} & {Fail} & {Fail} & {Fail} & {Fail} & {Fail} \\
    
    \textbf{Ours} & \bestbg{0.080} & \thirdbg{0.160} & \secondbg{0.336} & \bestbg{0.177} & \secondbg{1.080} & \bestbg{2.260} & \secondbg{0.338} & \thirdbg{2.900} & \secondbg{1.260} \\
    \bottomrule
    \end{tabular}
}

\end{table*}
\begin{table}[hbt!]
\centering

\caption{
\textbf{Mapping quality and system speed on Newer College and Oxford Spires datasets. } `Est.' indicates the use of online estimated poses. Metric units are cm for Acc, Com, C-L1, and \% for F-score. FPS denotes system FPS (Tracking + Mapping) or Mapping only.}

\label{tab:mapping_quality}

\setlength{\tabcolsep}{2.5pt} 

\resizebox{\linewidth}{!}{
\begin{tabular}{l|c|ccccc|ccccc|ccccc|ccccc}
\toprule

 & & \multicolumn{10}{c|}{\textbf{Newer College}} & \multicolumn{10}{c}{\textbf{Oxford Spires}} \\

\textbf{Method} & \textbf{Pose} & \multicolumn{5}{c|}{\textbf{Quad}} & \multicolumn{5}{c|}{\textbf{Math Institute}} & \multicolumn{5}{c|}{\textbf{Observatory}} & \multicolumn{5}{c}{\textbf{Keble College}} \\

 & & Acc$\downarrow$ & Com$\downarrow$ & C-L1$\downarrow$ & F$\uparrow$ & FPS$\uparrow$ & Acc$\downarrow$ & Com$\downarrow$ & C-L1$\downarrow$ & F$\uparrow$ & FPS$\uparrow$ & Acc$\downarrow$ & Com$\downarrow$ & C-L1$\downarrow$ & F$\uparrow$ & FPS$\uparrow$ & Acc$\downarrow$ & Com$\downarrow$ & C-L1$\downarrow$ & F$\uparrow$ & FPS$\uparrow$ \\

\midrule
Voxblox~\cite{oleynikova2017voxblox}         & GT   & 16.58 & 19.00 & 17.79 & 64.00 & \bestbg{66.10} & 12.98 & \thirdbg{15.02} & \thirdbg{14.00} & \thirdbg{77.23} & \bestbg{82.70} & \thirdbg{14.26} & \secondbg{17.47} & \thirdbg{15.86} & \thirdbg{71.73} & \bestbg{143.00} & \thirdbg{12.69} & \secondbg{14.97} & \thirdbg{13.83} & \thirdbg{77.29} & \bestbg{140.30} \\
VDBFusion~\cite{vizzo2022vdbfusion}       & GT   & 14.88 & \secondbg{14.09} & \thirdbg{14.49} & \thirdbg{76.27} & \secondbg{26.44} & \secondbg{9.04} & \bestbg{12.19} & \secondbg{10.61} & \secondbg{84.59} & \secondbg{45.40} & \secondbg{9.88} & \bestbg{11.97} & \bestbg{10.93} & \bestbg{83.68} & \secondbg{48.16} & \secondbg{9.11} & \bestbg{11.29} & \bestbg{10.20} & \bestbg{84.99} & \secondbg{43.69} \\
$N^3$-Mapping~\cite{song2024n}   & GT   & \secondbg{8.73} & \bestbg{12.33} & \bestbg{10.48} & \secondbg{85.66} & 0.68 & \bestbg{8.71} & \secondbg{12.42} & \bestbg{10.57} & \bestbg{84.83} & 0.68 & \bestbg{6.02} & \thirdbg{17.73} & \secondbg{11.87} & \secondbg{80.55} & 0.75 & \bestbg{4.82} & \thirdbg{18.25} & \secondbg{11.53} & \secondbg{80.87} & 0.79 \\
\midrule
PIN-SLAM~\cite{pan2024pin}        & Est. & 17.11 & 22.83 & 19.97 & 60.59 & 7.62 & 18.06 & 18.29 & 18.18 & 63.78 & 11.04 & 22.44 & 26.40 & 24.42 & 44.15 & 8.20 & 24.16 & 34.22 & 29.19 & 33.93 & 8.62 \\
Splat-LOAM~\cite{giacomini2025splat}      & Est. & \thirdbg{12.23} & 18.91 & 15.57 & 72.06 & 3.91 & \thirdbg{11.70} & 18.33 & 15.02 & 73.66 & 2.54 & 24.79 & 47.88 & 36.34 & 6.14 & 1.52 & 21.48 & 35.05 & 28.26 & 33.97 & 5.77 \\
\textbf{Ours}   & Est. & \bestbg{7.62} & \thirdbg{15.02} & \secondbg{11.32} & \bestbg{86.78} & \thirdbg{22.15} & 14.43 & 18.54 & 16.48 & 70.35 & \thirdbg{25.25} & 14.48 & 27.18 & 20.83 & 56.89 & \thirdbg{17.99} & 12.71 & 24.57 & 18.64 & 64.02 & \thirdbg{19.58} \\
\bottomrule
\end{tabular}
}

\end{table}

\subsection{Tracking}
We evaluate tracking performance on the KITTI Odometry~\cite{Geiger2012CVPR}, Oxford Spires~\cite{tao2025spires}, and Newer College~\cite{ramezani2020newer} datasets, as reported in \cref{tab:total_ate_evaluation}. Our method achieves competitive ATE for real-time dense Gaussian mapping, rather than uniformly best odometry. For Splat-LOAM, we used the official implementation with the provided \texttt{kitti.yaml} and no modification; a run is marked as failed when tracking diverges or ATE RMSE exceeds 50 m. Since we share the same loop-closure and pose-graph backend as KISS-SLAM, the differences mainly reflect the front-end scan-to-map tracking and map representation.

\subsection{System Speed and Quality of Reconstructed Map}

\begin{figure*}[t]
    \centering
    \includegraphics[width=\textwidth]{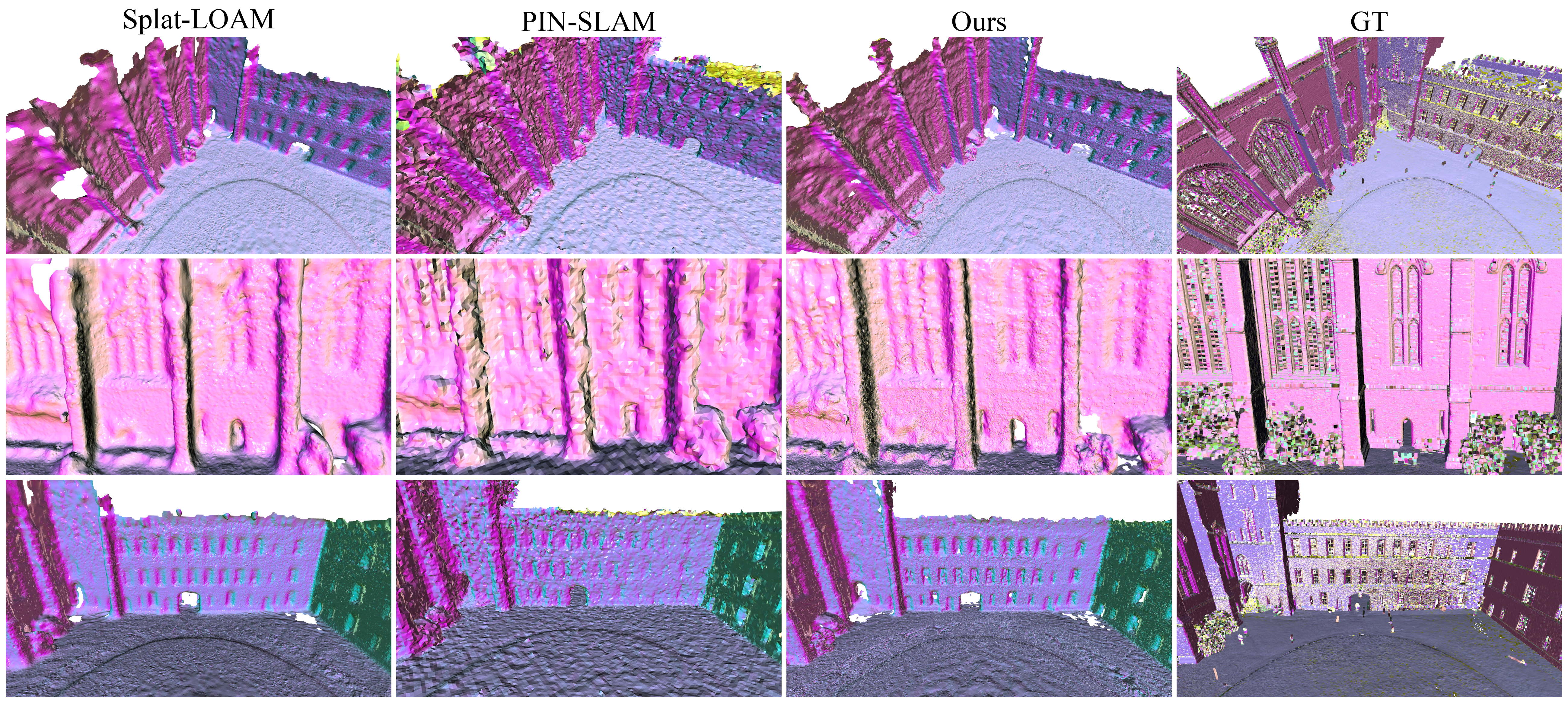}
    \caption{%
      \textbf{Qualitative mesh reconstruction comparison.} The figure shows reconstruction results for the quad-easy sequence using estimated poses from the Newer College dataset. Colors represent surface normals mapped to the RGB color space. Our method recovers geometry with high fidelity. PIN-SLAM shows high levels of noise, and Splat-LOAM shows areas of under-reconstruction and over-smoothing.
    }
    \label{fig:mesh}
    \vspace{-5pt}
\end{figure*}

To quantitatively evaluate the mapping quality, we convert the reconstructed GS maps into meshes and measure the Accuracy, Completeness, and F-score against the ground truth (GT) meshes. 
\cref{tab:mapping_quality} and Fig.~\ref{fig:mesh} present qualitative and quantitative results of the reconstructed scenes respectively. 

Among simultaneous tracking and mapping methods, our approach offers the best balance of quality, speed, and memory. Averaged over four sequences, it achieves an F-score of 69.51 at 21.24 FPS. This outperforms PIN-SLAM (50.61, 8.87 FPS) and Splat-LOAM (46.46, 3.44 FPS) by +18.90 and +23.05 F-score, with 2.39$\times$ and 6.18$\times$ speedups, respectively. Our map is also compact (7.23 MB on average), 61.6\% smaller than PIN-SLAM (18.83 MB), while delivering higher reconstruction quality. These results indicate that our map management is not mere compression: it preserves structurally important primitives while removing redundancies, enabling high-fidelity, real-time mapping on large-scale sequences.

\subsection{Ablation Study}
\label{sec:ablation}

\begin{table}[t!]
\centering
\caption{
\textbf{Ablation Results on Effect of Tracking-to-Mapping Information Transfer.}
We evaluate the impact of utilizing tracking-derived priors for the mapping process on the Quad and Keble College sequences. Tracking accuracy is measured with ATE RMSE [m], while map quality is assessed using distance metrics (Acc, Com, C-L1) in cm and F-score in \%.
}
\label{tab:ablation_t2m}
\resizebox{\linewidth}{!}{%
\begin{tabular}{@{} l | c c c c c | c c c c c @{}}
\toprule
 & \multicolumn{5}{c|}{\textbf{Quad}} & \multicolumn{5}{c}{\textbf{Keble College}} \\
\cmidrule(lr){2-6} \cmidrule(lr){7-11}
 & \textbf{ATE RMSE} & \multicolumn{4}{c|}{\textbf{Map Quality}} & \textbf{ATE RMSE} & \multicolumn{4}{c}{\textbf{Map Quality}} \\
\cmidrule(lr){3-6} \cmidrule(lr){8-11}
\textbf{Variant} & $\downarrow$ & \textbf{Acc}$\downarrow$ & \textbf{Com}$\downarrow$ & \textbf{C-L1}$\downarrow$ & \textbf{F-score}$\uparrow$ & $\downarrow$ & \textbf{Acc}$\downarrow$ & \textbf{Com}$\downarrow$ & \textbf{C-L1}$\downarrow$ & \textbf{F-score}$\uparrow$ \\
\midrule
A. w/o Init (all identity)   & 0.088 & 8.495 & 35.885 & 22.190 & 48.688 & 0.283 & 15.202 & 32.837 & 24.020 & 44.123 \\
B. w/o Init Range adap Scale & 0.087 & 8.016 & 15.254 & 11.635 & 85.996 & 0.479 & 16.451 & 28.666 & 22.558 & 51.582 \\
C. w/o Init Opacity          & 0.085 & 7.809 & 15.091 & 11.450 & 86.020 & 0.451 & 15.735 & 28.685 & 22.210 & 53.173 \\
D. w/o G-ICP Normal           & 0.099 & 8.040 & 15.160 & 11.600 & 85.899 & 0.325 & 15.678 & 27.179 & 21.428 & 55.522 \\
E. w/o Feat-aware            & 0.089 & 8.245 & 15.212 & 11.729 & 86.237 & 0.222 & 12.942 & 25.816 & 19.379 & 63.217 \\
F. Full                      & \textbf{0.079} & \textbf{7.624} & \textbf{15.020} & \textbf{11.320} & \textbf{86.776} & \textbf{0.212} & \textbf{12.713} & \textbf{24.568} & \textbf{18.640} & \textbf{64.024} \\
\bottomrule
\end{tabular}%
}

\end{table}

\noindent\textbf{Effect of Tracking-to-Mapping Information Transfer. }

\cref{tab:ablation_t2m} analyzes tracking-derived signals used in the mapping module.
In \cref{tab:ablation_t2m}.A, removing tracking covariances causes the largest degradation: without covariance-based orientation and scale priors, Gaussians initialized with identity rotations and unaligned scales lead to poor surface coverage and misoriented accumulation, destabilizing optimization under a fixed budget. Consequently, the F-score drops sharply on Quad, with clear deterioration in Com and C-L1.
Range-adaptive scale initialization is crucial for scenes with many distant returns; disabling it (\cref{tab:ablation_t2m}.B) reduces far-range coverage and degrades both tracking ATE and mapping quality (e.g., Keble College).
Removing opacity initialization (\cref{tab:ablation_t2m}.C) consistently lowers mapping quality, while its effect on tracking ATE varies by dataset, since ATE is also affected by correspondence distribution, visibility, and dynamic-object suppression.
Without the G-ICP normal loss (\cref{tab:ablation_t2m}.D), mapping quality changes little but tracking ATE worsens, indicating that normal supervision rapidly stabilizes tangent-plane orientations during mapping and yields more reliable target covariances/orientations for scan-to-map registration.
Finally, disabling feature-aware map management (\cref{tab:ablation_t2m}.E) has minimal short-term impact on mapping quality; because the control score mainly reallocates representation budget over time, we examine the long-horizon quality--resource trade-off in \cref{tab:ablation_manage_long}. Overall, using all components provides the best mapping quality and tracking accuracy.

\begin{table}[t!]
\centering
\caption{\textbf{Ablation Results on Map Management. }
We evaluate the impact of map management methods on map fidelity (Acc, Com, C-L1 in cm; F-score in \%) and resource consumption (\#GS, Map size, and Peak VRAM).}
\label{tab:ablation_manage_long}
\resizebox{\linewidth}{!}{%
\begin{tabular}{@{} l | c c c c c c c | c c c c c c c @{}}
\toprule
 & \multicolumn{7}{c|}{\textbf{Quad}} & \multicolumn{7}{c}{\textbf{Keble College}} \\
\cmidrule(lr){2-8} \cmidrule(lr){9-15}
 & \multicolumn{4}{c}{\textbf{Map Quality}} & \textbf{\#GS} & \textbf{Map} & \textbf{VRAM} & \multicolumn{4}{c}{\textbf{Map Quality}} & \textbf{\#GS} & \textbf{Map} & \textbf{VRAM} \\
\cmidrule(lr){2-5} \cmidrule(lr){6-8} \cmidrule(lr){9-12} \cmidrule(lr){13-15}
\textbf{Variant} & \textbf{Acc} $\downarrow$ & \textbf{Com} $\downarrow$ & \textbf{C-L1} $\downarrow$ & \textbf{F-score} $\uparrow$ & $\downarrow$ & $\downarrow$ & $\downarrow$ & \textbf{Acc} $\downarrow$ & \textbf{Com} $\downarrow$ & \textbf{C-L1} $\downarrow$ & \textbf{F-score} $\uparrow$ & $\downarrow$ & $\downarrow$ & $\downarrow$ \\
\midrule
Baseline (opacity prune) & 7.691 & 14.964 & 11.327 & 87.179 & 239,175 & 9.6MB & \textbf{2.9 GB} & 13.219 & 24.664 & 18.942 & \textbf{64.863} & 546,400 & 21.9MB & \textbf{4.1GB} \\
Split                    & \textbf{7.597} & \textbf{14.504} & \textbf{11.051} & \textbf{88.040} & 248,060 & 9.9MB & 4.8 GB & 15.718 & 27.220 & 21.469 & 56.681 & 591,567 & 23.7MB & 4.2GB \\
Plane Prune              & 7.708 & 15.100 & 11.404 & 86.521 & \textbf{119,285} & \textbf{4.8MB} & 4.5 GB & 14.694 & 26.163 & 20.429 & 59.327 & \textbf{341,435} & \textbf{13.7MB} & \textbf{4.1GB} \\
Split + Plane Prune      & 7.624 & 15.020 & 11.320 & 86.776 & 129,885 & 5.0MB & 4.5 GB & \textbf{12.713} & \textbf{24.568} & \textbf{18.640} & 64.024 &  344,473 & 13.8MB & 4.3GB\\
\bottomrule
\end{tabular}%
}
\end{table}
%
\begin{table}[t!]
\centering

\caption{
\textbf{Ablation Results on Effect of Mapping-to-Tracking Information Transfer.}
ATE RMSE is reported in meters. The \textit{Full (w/ feedback)} system achieves not only the lowest tracking drift but also the highest tracking FPS, demonstrating that optimized Gaussians serve as superior registration targets.
}
\label{tab:ablation_m2t}
\scriptsize
\setlength{\tabcolsep}{8pt}
\begin{tabular}{@{} l c c @{}}
\toprule
\textbf{Variant} & \textbf{ATE RMSE} $\downarrow$ & \textbf{Tracking FPS} $\uparrow$ \\
\midrule
w/o GS Opt. (frozen / no feedback) & 0.099 & 17.69 \\
w/o geometric weight               & 0.088  & 22.66 \\
w/o Conf/Filtering                 & 0.085 & 20.47 \\ 
Full (w/ feedback)                 & \textbf{0.079} & \textbf{23.70} \\
\bottomrule
\end{tabular}
\end{table}

\noindent\textbf{Effect of Mapping-to-Tracking Information Transfer. }

This ablation study evaluates the effect of feeding back the optimized GS map from the mapping module to the tracking process. 
In \cref{tab:ablation_m2t}, the w/o GS Opt. configuration disables GS map optimization (resulting in a frozen map) to compare the tracking accuracy and processing speed under an identical tracking pipeline.
Consequently, the Full system achieves a lower ATE RMSE and a higher tracking FPS compared to the frozen map baseline. This improvement indicates that the updated rotation, scale, and opacity of the target Gaussians from mapping optimization provide more stable covariance-based constraints during scan-to-map registration. As a result, the number of registration iterations and the ratio of failed correspondences decrease, yielding a distinct advantage in effective processing speed. Ultimately, this demonstrates that the coupling in our system is not a disruptive correction mechanism where "mapping interferes with tracking," but rather a purposefully designed feedback structure where the optimized GS map inherently serves as a superior registration target. 

\begin{table}[t!]
\centering
\vspace{-5pt}
\caption{\textbf{Ablation Results on Mapping Losses. }
We evaluate the impact of individual loss terms on tracking accuracy and mapping quality using the Quad and Keble College sequences. ATE RMSE is reported in meters, distance metrics (Acc, Com, C-L1) in cm, and F-score in \%.}
\label{tab:ablation_mapping_losses}

\resizebox{\linewidth}{!}{%
\begin{tabular}{@{} l | c c c c c | c c c c c @{}}
\toprule
 & \multicolumn{5}{c|}{\textbf{Quad}} & \multicolumn{5}{c}{\textbf{Keble College}} \\
\cmidrule(lr){2-6} \cmidrule(lr){7-11}
 & \textbf{ATE RMSE} & \multicolumn{4}{c|}{\textbf{Map Quality}} & \textbf{ATE RMSE} & \multicolumn{4}{c}{\textbf{Map Quality}} \\
\cmidrule(lr){3-6} \cmidrule(lr){8-11}
\textbf{Variant} & $\downarrow$ & \textbf{Acc} $\downarrow$ & \textbf{Com} $\downarrow$ & \textbf{C-L1} $\downarrow$ & \textbf{F-score} $\uparrow$ & $\downarrow$ & \textbf{Acc} $\downarrow$ & \textbf{Com} $\downarrow$ & \textbf{C-L1} $\downarrow$ & \textbf{F-score} $\uparrow$ \\
\midrule
w/o G-ICP Normal   & 0.097 & 8.119 & 15.247 & 11.683 & 85.844 & 0.346 & 16.274 & 27.222 & 22.248 & 54.206 \\
w/o Scale         & 0.086 & 7.788 & 15.859 & 11.823 & 85.732 & 0.262 & 21.017 & 35.305 & 28.161 & 33.924 \\
w/o Opacity       & 0.084 & 8.681 & 16.230 & 12.456 & 83.217 & 0.328 & 16.283 & 27.678 & 21.980 & 52.893 \\
Full              & \textbf{0.079} & \textbf{7.624} & \textbf{15.020} & \textbf{11.320} & \textbf{86.776} & \textbf{0.212} & \textbf{12.713} & \textbf{24.568} &\textbf{18.640} & \textbf{64.024} \\
\bottomrule
\end{tabular}%
}

\end{table}

\noindent\textbf{Mapping Loss.}
\label{sec:ablation_losses}

\cref{tab:ablation_mapping_losses} analyzes the impact of each term in the mapping objective on map quality and tracking performance. The Full configuration utilizes all loss terms.
\textit{w/o G-ICP Normal} removes the normal supervision transferred from the tracking process, leading to a distinct degradation in ATE across both sequences. We attribute this to the fact that the depth consistency loss alone provides weak local surface orientation constraints. Consequently, the tangent plane orientations of target surfels are inadequately reflected, biasing the optimization towards simply generating overly smooth surfaces. This lack of geometric detail ultimately disadvantages scan-to-map registration.
\textit{w/o Opacity} ablates the term that induces opacity from valid observation pixels, resulting in the most severe degradation in map quality. Without this term, uncertain observations are not sufficiently suppressed, or conversely, coverage in necessary areas weakens. This leads to an increase in holes and ghosting artifacts, simultaneously deteriorating both precision and completeness.
\textit{w/o Scale} disables scale regularization. Under this setting, certain Gaussians can become excessively large, over-covering surrounding surfaces. This phenomenon causes a sharp decline in map quality metrics, particularly in the Keble College sequence.
In summary, the proposed loss combination simultaneously enhances depth-based registration (geometry), orientation constraints (normal), and representation stability (opacity and scale), continuously providing a highly favorable registration target for the tracking process.

\begin{figure*}[t!]
    \centering
    \includegraphics[width=\textwidth]{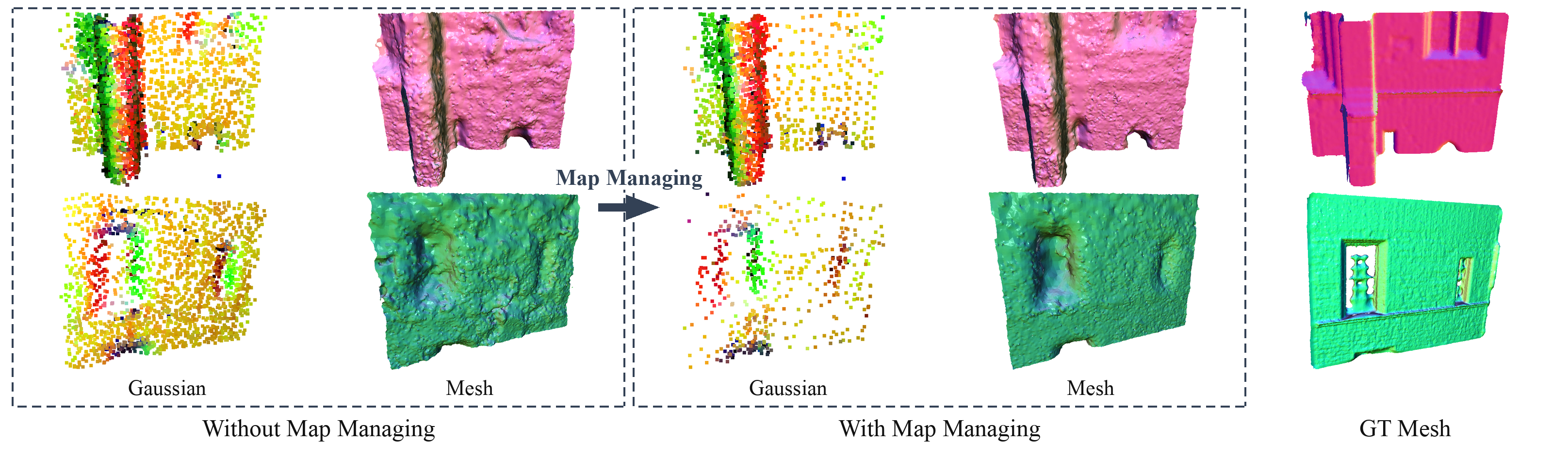}
    \caption{\textbf{Geometry-aware map budget control driven by covariance-derived control scores. }The map remains compact in large planar areas while preserving fine structures.}
    \label{fig:map_managing_result}
\end{figure*}
%
\noindent\textbf{Gaussian Map Management. }
\label{sec:ablation_management}

\cref{tab:ablation_manage_long} evaluates how our map-management strategy affects the long-horizon trade-off between map quality and resource usage (number of Gaussians). 
\textit{Baseline} applies only opacity-based pruning. Plane Prune performs control-score-based planar compression via a cover-and-prune scheme, while Split reinforces complex structures through splitting. Split + Plane Prune is our final configuration combining both.

Plane Prune removes redundant Gaussians in large planar regions, substantially reducing the total Gaussian count and map size, at the cost of a slight quality drop due to losing some fine details. 
Conversely, Split increases the number of Gaussians but improves representational capacity in complex areas, often boosting the F-score. 
Using both mechanisms strikes the best balance. 
Planar compression keeps map growth in check, while splitting restores fine detail where it matters, leading to the strongest quality--resource trade-off and more reliable registration targets over long sequences.
\cref{fig:map_managing_result} visually confirms this behavior, showing a sparse distribution of Gaussians in planar regions and a dense concentration within complex geometric structures.

We further analyze scalability on KITTI sequence 00. Since KITTI does not provide ground-truth meshes, we report only resource consumption. 
With our management strategy, the number of Gaussians decreases from 2,482,505 to 1,510,033, and the persistent map storage drops from 61.3\,MB to 42.1\,MB, demonstrating a substantial reduction in stored map size in a practical large-scale environment.

\section{Conclusion}

We presented a real-time LiDAR-based Gaussian Splatting SLAM system for large-scale outdoor environments. The core idea is to reuse local covariance information across tracking and mapping: tracking covariances provide geometric priors for Gaussian initialization and map management, while mapping-refined Gaussian geometry provides covariance-aware targets for scan-to-map registration. Together with the proposed control-score-based pruning and densification strategy, this coupling allocates map capacity to structurally informative regions while reducing redundant Gaussians in simpler areas. As a result, the system achieves real-time performance with compact persistent map storage and competitive peak VRAM in full-system comparisons. These results suggest that covariance-coupled tracking and map management are effective design choices for scalable LiDAR Gaussian Splatting SLAM.

\paragraph{Limitations.}
Our system relies on local geometric estimates from G-ICP. When the scene has weak geometric constraints, sparse returns, vegetation, or many dynamic objects, the estimated covariances and normals can become less reliable and may affect tracking and mapping. Several parameters, such as voxel resolution, Gaussian scale bounds, and keyframe thresholds, are tied to sensor resolution and scene scale. Loop closure also updates associated Gaussians in a keyframe-wise rigid manner rather than through full global Gaussian re-optimization, which may leave local inconsistencies after large corrections. KITTI is useful for large-scale evaluation but has known ground-truth limitations.

\bibliographystyle{splncs04}
\bibliography{main}

\begin{thebibliography}{10}
\providecommand{\url}[1]{\texttt{#1}}
\providecommand{\urlprefix}{URL }
\providecommand{\doi}[1]{https://doi.org/#1}

\bibitem{behley2018efficient}
Behley, J., Stachniss, C.: Efficient surfel-based slam using 3d laser range
  data in urban environments. In: Robotics: science and systems. vol.~2018,
  p.~59 (2018)

\bibitem{chen2022direct}
Chen, K., Lopez, B.T., Agha-mohammadi, A.a., Mehta, A.: Direct lidar odometry:
  Fast localization with dense point clouds. IEEE Robotics and Automation
  Letters  \textbf{7}(2),  2000--2007 (2022)

\bibitem{chen2019suma++}
Chen, X., Milioto, A., Palazzolo, E., Giguere, P., Behley, J., Stachniss, C.:
  Suma++: Efficient lidar-based semantic slam. In: 2019 IEEE/RSJ International
  Conference on Intelligent Robots and Systems (IROS). pp. 4530--4537. IEEE
  (2019)

\bibitem{dellenbach2022ct}
Dellenbach, P., Deschaud, J.E., Jacquet, B., Goulette, F.: Ct-icp: Real-time
  elastic lidar odometry with loop closure. In: 2022 international conference
  on robotics and automation (ICRA). pp. 5580--5586. IEEE (2022)

\bibitem{deng2023nerf}
Deng, J., Wu, Q., Chen, X., Xia, S., Sun, Z., Liu, G., Yu, W., Pei, L.:
  Nerf-loam: Neural implicit representation for large-scale incremental lidar
  odometry and mapping. In: Proceedings of the IEEE/CVF International
  Conference on Computer Vision. pp. 8218--8227 (2023)

\bibitem{Geiger2012CVPR}
Geiger, A., Lenz, P., Urtasun, R.: Are we ready for autonomous driving? the
  kitti vision benchmark suite. In: Conference on Computer Vision and Pattern
  Recognition (CVPR) (2012)

\bibitem{giacomini2025splat}
Giacomini, E., Di~Giammarino, L., De~Rebotti, L., Grisetti, G., Oswald, M.R.:
  Splat-loam: Gaussian splatting lidar odometry and mapping. In: Proceedings of
  the IEEE/CVF International Conference on Computer Vision. pp. 27630--27639
  (2025)

\bibitem{kiss2025iros}
Guadagnino, T., Mersch, B., Gupta, S., Vizzo, I., Grisetti, G., Stachniss, C.:
  {KISS-SLAM: A Simple, Robust, and Accurate 3D LiDAR SLAM System With Enhanced
  Generalization Capabilities}. In: 2025 IEEE/RSJ International Conference on
  Intelligent Robots and Systems (IROS). pp. 5363--5370 (2025).
  \doi{10.1109/IROS60139.2025.11246613}

\bibitem{gupta2024icra}
Gupta, S., Guadagnino, T., Mersch, B., Vizzo, I., Stachniss, C.: {Effectively
  Detecting Loop Closures using Point Cloud Density Maps}. In: IEEE
  International Conference on Robotics and Automation (ICRA) (2024)

\bibitem{ha2024rgbd}
Ha, S., Yeon, J., Yu, H.: Rgbd gs-icp slam. In: European conference on computer
  vision. pp. 180--197. Springer (2024)

\bibitem{he2023point}
He, D., Xu, W., Chen, N., Kong, F., Yuan, C., Zhang, F.: Point-lio: robust
  high-bandwidth light detection and ranging inertial odometry. Advanced
  Intelligent Systems  \textbf{5}(7),  2200459 (2023)

\bibitem{7989597}
Hsiao, M., Westman, E., Zhang, G., Kaess, M.: Keyframe-based dense planar slam.
  In: 2017 IEEE International Conference on Robotics and Automation (ICRA). pp.
  5110--5117 (2017). \doi{10.1109/ICRA.2017.7989597}

\bibitem{huang20242d}
Huang, B., Yu, Z., Chen, A., Geiger, A., Gao, S.: 2d gaussian splatting for
  geometrically accurate radiance fields. In: ACM SIGGRAPH 2024 conference
  papers. pp. 1--11 (2024)

\bibitem{huang2024photo}
Huang, H., Li, L., Cheng, H., Yeung, S.K.: Photo-slam: Real-time simultaneous
  localization and photorealistic mapping for monocular stereo and rgb-d
  cameras. In: Proceedings of the IEEE/CVF Conference on Computer Vision and
  Pattern Recognition. pp. 21584--21593 (2024)

\bibitem{keetha2024splatam}
Keetha, N., Karhade, J., Jatavallabhula, K.M., Yang, G., Scherer, S., Ramanan,
  D., Luiten, J.: Splatam: Splat track \& map 3d gaussians for dense rgb-d
  slam. In: Proceedings of the IEEE/CVF conference on computer vision and
  pattern recognition. pp. 21357--21366 (2024)

\bibitem{kerbl3Dgaussians}
Kerbl, B., Kopanas, G., Leimk{\"u}hler, T., Drettakis, G.: 3d gaussian
  splatting for real-time radiance field rendering. ACM Transactions on
  Graphics  \textbf{42}(4) (July 2023),
  \url{https://repo-sam.inria.fr/fungraph/3d-gaussian-splatting/}

\bibitem{lang2025gaussian}
Lang, X., Lv, J., Tang, K., Li, L., Huang, J., Liu, L., Liu, Y., Zuo, X.:
  Gaussian-lic2: Lidar-inertial-camera gaussian splatting slam. arXiv preprint
  arXiv:2507.04004  (2025)

\bibitem{liu2025gs}
Liu, J., Wan, Y., Wang, B., Zheng, C., Lin, J., Zhang, F.: Gs-sdf:
  Lidar-augmented gaussian splatting and neural sdf for geometrically
  consistent rendering and reconstruction. In: 2025 IEEE/RSJ International
  Conference on Intelligent Robots and Systems (IROS). pp. 19391--19398. IEEE
  (2025)

\bibitem{matsuki2024gaussian}
Matsuki, H., Murai, R., Kelly, P.H., Davison, A.J.: Gaussian splatting slam.
  In: Proceedings of the IEEE/CVF conference on computer vision and pattern
  recognition. pp. 18039--18048 (2024)

\bibitem{oleynikova2017voxblox}
Oleynikova, H., Taylor, Z., Fehr, M., Siegwart, R., Nieto, J.: Voxblox:
  Incremental 3d euclidean signed distance fields for on-board mav planning.
  In: IEEE/RSJ International Conference on Intelligent Robots and Systems
  (IROS) (2017)

\bibitem{pak2025g2s}
Pak, G., Cho, H.M., Kim, E.: G2s-icp slam: Geometry-aware gaussian splatting
  icp slam. arXiv preprint arXiv:2507.18344  (2025)

\bibitem{pan2025pings}
Pan, Y., Zhong, X., Jin, L., Wiesmann, L., Popovi{\'c}, M., Behley, J.,
  Stachniss, C.: Pings: Gaussian splatting meets distance fields within a
  point-based implicit neural map. arXiv preprint arXiv:2502.05752  (2025)

\bibitem{pan2024pin}
Pan, Y., Zhong, X., Wiesmann, L., Posewsky, T., Behley, J., Stachniss, C.:
  Pin-slam: Lidar slam using a point-based implicit neural representation for
  achieving global map consistency. IEEE Transactions on Robotics  \textbf{40},
   4045--4064 (2024)

\bibitem{ramezani2020newer}
Ramezani, M., Wang, Y., Camurri, M., Wisth, D., Mattamala, M., Fallon, M.: The
  newer college dataset: Handheld lidar, inertial and vision with ground truth.
  In: 2020 IEEE/RSJ International Conference on Intelligent Robots and Systems
  (IROS). pp. 4353--4360 (2020). \doi{10.1109/IROS45743.2020.9340849}

\bibitem{segal2009generalized}
Segal, A., Haehnel, D., Thrun, S., et~al.: Generalized-icp. In: Robotics:
  science and systems. vol.~2, p.~435. Seattle, WA (2009)

\bibitem{shan2018lego}
Shan, T., Englot, B.: Lego-loam: Lightweight and ground-optimized lidar
  odometry and mapping on variable terrain. In: 2018 IEEE/RSJ international
  conference on intelligent robots and systems (IROS). pp. 4758--4765. IEEE
  (2018)

\bibitem{shan2020lio}
Shan, T., Englot, B., Meyers, D., Wang, W., Ratti, C., Rus, D.: Lio-sam:
  Tightly-coupled lidar inertial odometry via smoothing and mapping. In: 2020
  IEEE/RSJ international conference on intelligent robots and systems (IROS).
  pp. 5135--5142. IEEE (2020)

\bibitem{song2024n}
Song, S., Zhao, J., Huang, K., Lin, J., Ye, C., Feng, T.: {$N^3$-Mapping}:
  Normal guided neural non-projective signed distance fields for large-scale 3d
  mapping. IEEE Robotics and Automation Letters  \textbf{9}(6),  5935--5942
  (2024)

\bibitem{tao2025spires}
Tao, Y., Mu{\~n}oz-Ba{\~n}{\'o}n, M.{\'A}., Zhang, L., Wang, J., Fu, L.F.T.,
  Fallon, M.: The oxford spires dataset: Benchmarking large-scale lidar-visual
  localisation, reconstruction and radiance field methods. International
  Journal of Robotics Research  (2025)

\bibitem{turki2022mega}
Turki, H., Ramanan, D., Satyanarayanan, M.: Mega-nerf: Scalable construction of
  large-scale nerfs for virtual fly-throughs. In: Proceedings of the IEEE/CVF
  conference on computer vision and pattern recognition. pp. 12922--12931
  (2022)

\bibitem{vizzo2022vdbfusion}
Vizzo, I., Guadagnino, T., Behley, J., Stachniss, C.: Vdbfusion: Flexible and
  efficient tsdf integration of range sensor data. Sensors  \textbf{22}(3),
  ~1296 (2022)

\bibitem{xiao2024liv}
Xiao, R., Liu, W., Chen, Y., Hu, L.: Liv-gs: Lidar-vision integration for 3d
  gaussian splatting slam in outdoor environments. IEEE Robotics and Automation
  Letters  \textbf{10}(1),  421--428 (2024)

\bibitem{xie2025gs}
Xie, Y., Huang, Z., Wu, J., Ma, J.: Gs-livm: Real-time photo-realistic
  lidar-inertial-visual mapping with gaussian splatting. In: Proceedings of the
  IEEE/CVF International Conference on Computer Vision. pp. 26869--26878 (2025)

\bibitem{xu2022fast}
Xu, W., Cai, Y., He, D., Lin, J., Zhang, F.: Fast-lio2: Fast direct
  lidar-inertial odometry. IEEE Transactions on Robotics  \textbf{38}(4),
  2053--2073 (2022)

\bibitem{xu2021fast}
Xu, W., Zhang, F.: Fast-lio: A fast, robust lidar-inertial odometry package by
  tightly-coupled iterated kalman filter. IEEE Robotics and Automation Letters
  \textbf{6}(2),  3317--3324 (2021)

\bibitem{yan2024gs}
Yan, C., Qu, D., Xu, D., Zhao, B., Wang, Z., Wang, D., Li, X.: Gs-slam: Dense
  visual slam with 3d gaussian splatting. In: Proceedings of the IEEE/CVF
  conference on computer vision and pattern recognition. pp. 19595--19604
  (2024)

\bibitem{yang2022vox}
Yang, X., Li, H., Zhai, H., Ming, Y., Liu, Y., Zhang, G.: Vox-fusion: Dense
  tracking and mapping with voxel-based neural implicit representation. In:
  2022 IEEE International Symposium on Mixed and Augmented Reality (ISMAR). pp.
  499--507. IEEE (2022)

\bibitem{yeon2026gso}
Yeon, J., Ha, S., Yu, H.: Gso-slam: Bidirectionally coupled gaussian splatting
  and direct visual odometry. IEEE Robotics and Automation Letters  (2026)

\bibitem{zhang2014loam}
Zhang, J., Singh, S., et~al.: Loam: Lidar odometry and mapping in real-time.
  In: Robotics: Science and systems. vol.~2, pp.~1--9. Berkeley, CA (2014)

\bibitem{zhong2023shine}
Zhong, X., Pan, Y., Behley, J., Stachniss, C.: Shine-mapping: Large-scale 3d
  mapping using sparse hierarchical implicit neural representations. In: 2023
  IEEE International Conference on Robotics and Automation (ICRA). pp.
  8371--8377. IEEE (2023)

\bibitem{zhou2021pi}
Zhou, L., Wang, S., Kaess, M.: $\pi$-lsam: Lidar smoothing and mapping with
  planes. In: 2021 IEEE international conference on robotics and automation
  (ICRA). pp. 5751--5757. IEEE (2021)

\end{thebibliography}
\end{document}